\documentclass[authoryear,preprint,review,12pt]{elsarticle}

\usepackage{amsmath}
\usepackage{amssymb}
\usepackage{natbib}
\setcitestyle{square, numbers, sort&compress}
\usepackage[backref]{hyperref}
\usepackage{multirow}
\usepackage{subfig}
\usepackage{subcaption}

\usepackage{times}
\usepackage{soul}
\usepackage{url}
\usepackage{graphicx}
\usepackage{amsmath}
\usepackage{amsthm}
\usepackage{booktabs}
\usepackage{algorithm}
\usepackage{algorithmic}
\usepackage{makecell}
\usepackage[switch]{lineno}

\usepackage{color}
\usepackage{ulem}
\usepackage{natbib}
\setcitestyle{numbers,square}

\journal{Pattern Recognition}

\begin{document}

\begin{frontmatter}



\title{SiGNN: A Spike-induced Graph Neural Network for Dynamic Graph Representation Learning}

\author[a]{Dong Chen}
\ead{23120286@bjtu.edu.cn} 

\author[a]{Shuai Zheng}
\ead{zs1997@bjtu.edu.cn}
 
\author[a]{Muhao Xu}
\ead{muhx1998@bjtu.edu.cn}

\author[a]{Zhenfeng Zhu\corref{cor1}}
\ead{zhfzhu@bjtu.edu.cn}

\author[a]{Yao Zhao}
\ead{yzhao@bjtu.edu.cn}

\cortext[cor1]{Corresponding author}

\affiliation[a]{organization={Institute of Information Science, Beijing Jiaotong University},
            city={Beijing},
            postcode={100044}, 
            country={China}}

\begin{abstract}
In the domain of dynamic graph representation learning (DGRL), the efficient and comprehensive capture of temporal evolution within real-world networks is crucial. Spiking Neural Networks (SNNs), known as their temporal dynamics and low-power characteristic, offer an efficient solution  for temporal processing in DGRL task. However, owing to the spike-based information encoding mechanism of SNNs, existing DGRL methods employed SNNs face limitations in their representational capacity. Given this issue, we propose a novel framework named  \textbf{S}pike-\textbf{i}nduced \textbf{G}raph \textbf{N}eural \textbf{N}etwork (\textbf{SiGNN}) for learning enhanced spatial-temporal representations on dynamic graphs. In detail, a harmonious integration of SNNs and GNNs is achieved through an innovative \textbf{T}emporal \textbf{A}ctivation (\textbf{TA}) mechanism. Benefiting from the TA mechanism, SiGNN not only effectively exploits the temporal dynamics of SNNs but also adeptly circumvents the representational constraints imposed by the binary nature of spikes.  Furthermore, leveraging the inherent adaptability of SNNs, we explore an in-depth analysis of the evolutionary patterns within dynamic graphs across multiple time granularities. This approach facilitates the acquisition of a multiscale temporal node representation. Extensive experiments on various real-world dynamic graph datasets demonstrate the superior performance of SiGNN in the node classification task.
\end{abstract}



\end{frontmatter}


\section{Introduction}
\label{sec:intro}

Dynamic graph-structured data is common in real-world scenarios, including evolving relationships in social networks \cite{greene2010tracking}, fluctuating traffic flows in transportation networks \cite{gcn_7_kipf,WENG2023109670}, and ever-changing protein interactions \cite{ou2014detecting} in the field of biology. Efficiently and comprehensively capturing these evolutionary dynamics is crucial for a deeper understanding of how networks or systems change over time. In the field of Dynamic Graph Representation Learning (DGRL), a common strategy is to combine temporal models with Graph Neural Networks (GNNs), which are known for their powerful capability to handle graph data \cite{kipf2016semi}. Recurrent Neural Networks (RNNs) ~\cite{cho2014learning}, renowned for their proficiency in handling sequential data, have found application in DGRL task. However, they present challenges, including diminished computational efficiency \cite{sak2014long}, particularly in scenarios involving large-scale and lengthy sequences dynamic graphs. In light of these limitations, Spiking Neural Networks (SNNs)~\cite{gerstner2002spiking} have emerged as a promising alternative, providing low-power solutions to the limitations encountered by RNNs.

\begin{figure}
  \centering
  \includegraphics[width=10cm]{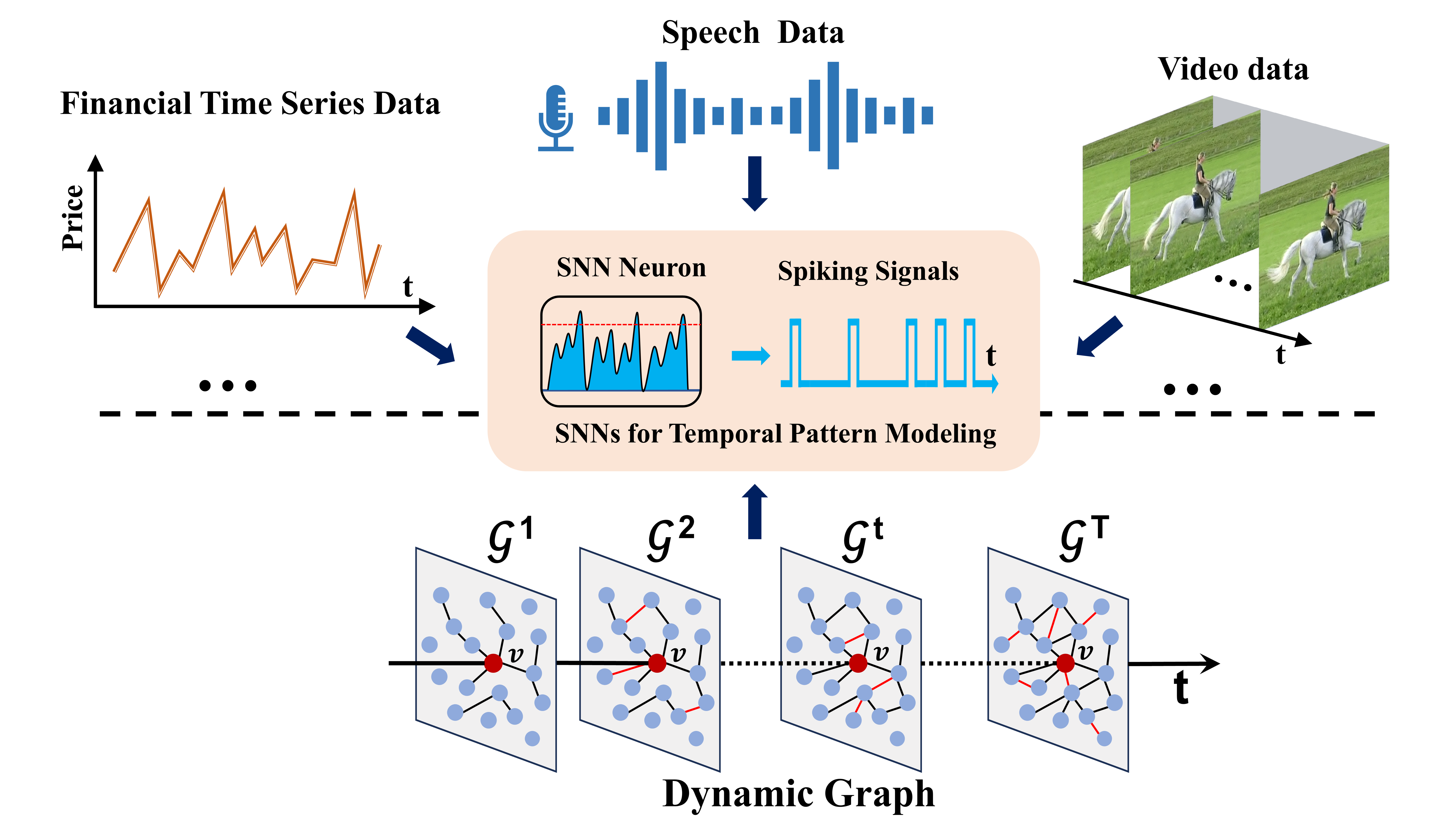}
  \caption{SNNs are widely employed for efficiently processing time-series data. Given that dynamic graph data also exhibit temporal dependencies, SNNs hold significant potential for capturing the temporal dynamics in evolving graphs.}\label{intro}
\end{figure}

SNNs encode information in patterns of spikes and each spiking neuron manifests rich dynamic behaviors \cite{deng2020rethinking}. Specifically, besides the spatial propagation, the current state of neurons is tightly affected by their historical activity in temporal domain. This nature introduces a temporal aspect to the processing of information, as the timing and sequence of spikes are critical for conveying time-related data. Furthermore, the internal computations of SNN neurons are relatively efficient, primarily involving linear integration and conditional judgment, devoid of complex nonlinear operations or matrix multiplication. Figure \ref{intro} demonstrates the utilization of SNNs in processing time series data across diverse domains, including financial data \cite{reid2014financial}, speech data \cite{wu2020deep}, video data \cite{zhu2022event}. Given that dynamic graphs manifest temporal dependencies similar to sequential data, SNNs hold the capacity to discern and capture the evolutionary patterns inherent in dynamic graphs.

Despite a few efforts to harness the temporal processing capacity of SNNs for learning spatial-temporal representations in dynamic graphs, they continue to confront certain challenges. Previous studies have predominantly focused on explicit spike-based methods, in which spikes obtained from SNNs are used directly as hidden features of nodes within the GNNs. Consequently, the issue of the reduced model's representational capacity \cite{han2020rmp} due to the binarization of features in SNNs is introduced, resulting in a diminished quality of learnt representations. Moreover, the majority of DGRL methods analyze dynamic graphs at a singular time scale, lacking insights across multiple temporal dimensions. This constraint hinders their ability to comprehensively understand the intricate  evolutionary dynamics within graphs.

To address these challenges, we propose the \textbf{S}pike-\textbf{i}nduced \textbf{G}raph \textbf{N}eural \textbf{N}etwork (\textbf{SiGNN}), an innovative framework that utilizes SNNs as temporal architectures while still maintaining exceptional representational capacity. By adopting a novel \textbf{T}emporal \textbf{A}ctivation (\textbf{TA}) mechanism, SiGNN seamlessly integrates the temporal dynamics of SNNs with the powerful capabilities of GNNs in processing graph data. Instead of treating spikes as features, the TA mechanism ingeniously utilizes spikes as activation signals to modulate feature propagation within GNNs temporally, thereby circumventing the binarization of features. Furthermore, we extend the application of the TA mechanism across Multiple Time Granularities (MTG). This approach endows SiGNN with the ability to intricately discern the evolutionary patterns of dynamic graphs from perspectives of various time resolutions. The contributions of our work are summarized as follows:
\begin{itemize}
   \item We propose a novel framework named SiGNN for learning enhanced spatial-temporal node representations in dynamic graphs. SiGNN intricately exploits the temporal processing capability of SNNs to capture the multiscale dynamics of evolving graphs. 
   \item Considering the limited representational capacity of binary spike signals, an innovative TA mechanism is introduced to harmoniously integrate the temporal dynamics of SNNs into static GNNs. 
   \item 
    To comprehensively capture the temporal evolution of dynamic graphs, our SiGNN implementation incorporates analyses at various time granularities, enabling the learning of node representations enriched with insights spanning multiple temporal scales.
   \item Extensive experiments demonstrate the outstanding performance of SiGNN in node classification tasks on real-world dynamic graph datasets, outperforming state-of-the-art approaches. 
\end{itemize}

\section{Related Work}
\label{sec:RelatedWork}

\subsection{Dynamic Graph Representation Learning}
In recent years, the field of learning representations for dynamic graphs has attracted considerable attention \cite{kazemi2020representation}. Unlike methods for static graphs, approaches designed for dynamic graphs necessitate additional consideration of the time dimension \cite{zuo2018embedding}. A prevalent approach for learning representations on dynamic graph involves the integration of RNNs \cite{shi2021gaen}, primarily due to their robust ability to handle sequential data. For instance, JODIE \cite{kumar2019predicting} and TGN \cite{rossi2020temporal} employ RNN units to update node memory and hidden states, thereby facilitating a coherent representation of interaction events. In a similar vein, EvolveGCN \cite{pareja2020evolvegcn} employs RNNs to continuously update GNN parameters  at each time step, thereby enhancing the model’s adaptability to time-varying changes. Furthermore, self-attention mechanisms have proven instrumental in capturing temporal patterns in dynamic graph data \cite{trivedi2019dyrep}.  For example, TGAT \cite{xu2020inductive} implements a temporal graph attention layer to efficiently aggregate features from temporal-topological neighborhoods. DySAT \cite{sankar2020dysat} introduces a technique for computing node representations using joint self-attention across spatial and temporal dimensions.

While methods based on RNNs or self-attention mechanisms are effective, they can be computationally expensive and resource-intensive. To address this challenge, some researchers have proposed methods based on SNNs to achieve more efficient learning of dynamic graph representations.

\subsection{SNN-based Graph Representation Learning}
SNNs are a class of brain-inspired with distinctive properties such as low power consumption and temporal dynamic \cite{kim2020spiking, bu2022optimized}. Endeavors have been undertaken to employ SNNs within the realm of graph representation learning. For example, SpikeE \cite{dold2021spike} is a model that leverages spike-based embeddings for the purpose of handling multirelational graph data. SpikingGCN \cite{zhu2022spiking} is a groundbreaking SNN tailored for node classification within graph data. It effectively integrates the  embeddings derived from GNNs with the intrinsic biofidelity characteristics of SNNs. SpikeGCL \cite{li2023graph} is a framework that combines SNNs with graph contrastive learning method for learning binarized representations that are both biologically plausible and compact.

One key characteristic of SNNs is their ability to capture temporal dynamics effectively. This unique feature enables SNNs to excel at representing and processing dynamic graphs by precisely encoding spike timing. Consequently, a few methods for dynamic graph representation learning based on SNNs have emerged in recent research. An exemple is SpikeNet \cite{li2023scaling}, a scalable framework that leverages the Leaky Integrate and Fire (LIF) model \cite{gerstner2014neuronal} to learn binarized spatial-temporal node representations on dynamic graphs. However, these methods are diminished by the limited representational capacity of SNNs. Further exploration is needed to fully exploit the temporal dynamics of SNNs for dynamic graph representation learning.

\section{Preliminary}\label{sec:problemFormulation}
\subsection{Problem Definition}

In this work, we focus on the task of dynamic graph representation learning in a discrete-time scenario. A discrete dynamic graph is a sequence of graph snapshots at $T$ different time steps, denoted by $\mathcal{G}=\{\mathcal{G}^1, \mathcal{G}^2,..., \mathcal{G}^T \}$. The snapshot at time $t$ is defined as $\mathcal{G}^t = (\mathcal{V}, \mathcal{E}^t, \mathbf{X}^t, \mathcal{A}^t)$, where $\mathcal{V}$ is a set of $n$ nodes, $\mathcal{E}^t$ is the set of edges among nodes, and $\mathbf{X}^t \in \mathbb{R}^{n \times d}$ represents the features of nodes. The topology structure of snapshot $\mathcal{G}^t$ could be described by the adjacency matrix $\mathcal{A}^t \in \mathbb{R}^{n \times n}$ with $\mathcal{A}^t_{i,j}=1$ if $(i,j)\in \mathcal{E}^t$ or 0 otherwise.  We assume that for any time snapshot of the $\mathcal{G}$, we use a common set of nodes and treat non-existent nodes as having zero degree. For each node $v \in \mathcal{V}$, we denote its neighborhood set at time $t$ by $N^t_v$.
Our goal is to learn node embeddings $Z$ for all nodes appeared in a discrete dynamic graph $\mathcal{G}$. These embeddings should encompass the temporal evolution dynamics of nodes, which can be subsequently leveraged for various downstream tasks.

\subsection{Leaky Integrate-and-Fire Model}

The Leaky Integrate-and-Fire (LIF) model, a quintessential neuron model in Spiking Neural Networks (SNNs), simulates neuronal behavior by accumulating membrane voltage over time, akin to a charging capacitor, until it reaches a specific threshold $V_{th}$. Upon surpassing this threshold, the neuron fires a spike action potential and resets its membrane voltage to a constant value $V_{reset}$. Additionally, this model includes a leak current term to simulate how the neuron's voltage slowly returns to the baseline level. LIF model assumes that the neuron's activity can be described by a differential equation as shown in \eqref{LIF}, which represents the change in the neuron's membrane voltage.

\begin{equation}\label{LIF}
\tau \frac{dV}{dt} = -(V - V_{\text{reset}}) + I
\end{equation}
where $\tau$ is a membrane-related factor to control how fast the membrane voltage decays. $I$ and $V$ are the input current and membrane voltage value of the LIF neuron. To better describe the neuron behaviors and guarantee computational availability, the differential equation can be converted to an iterative expression \cite{fang2021incorporating} as follow:

\begin{equation}\label{LIFi}
V^t = V^{t-1} + \frac{1}{\tau} (-(V^{t-1} - V_{reset}) + I^t)
\end{equation}
where  illustrates the computation of the membrane voltage $V^t$ at time $t$ based on the previous time's voltage $V^{t-1}$ and the input current $I^t$. It effectively captures the dynamic process of how the LIF neuron's voltage changes over time in response to input current.

\begin{figure*}[t]
  \centering
   \includegraphics[width=1.0\linewidth]{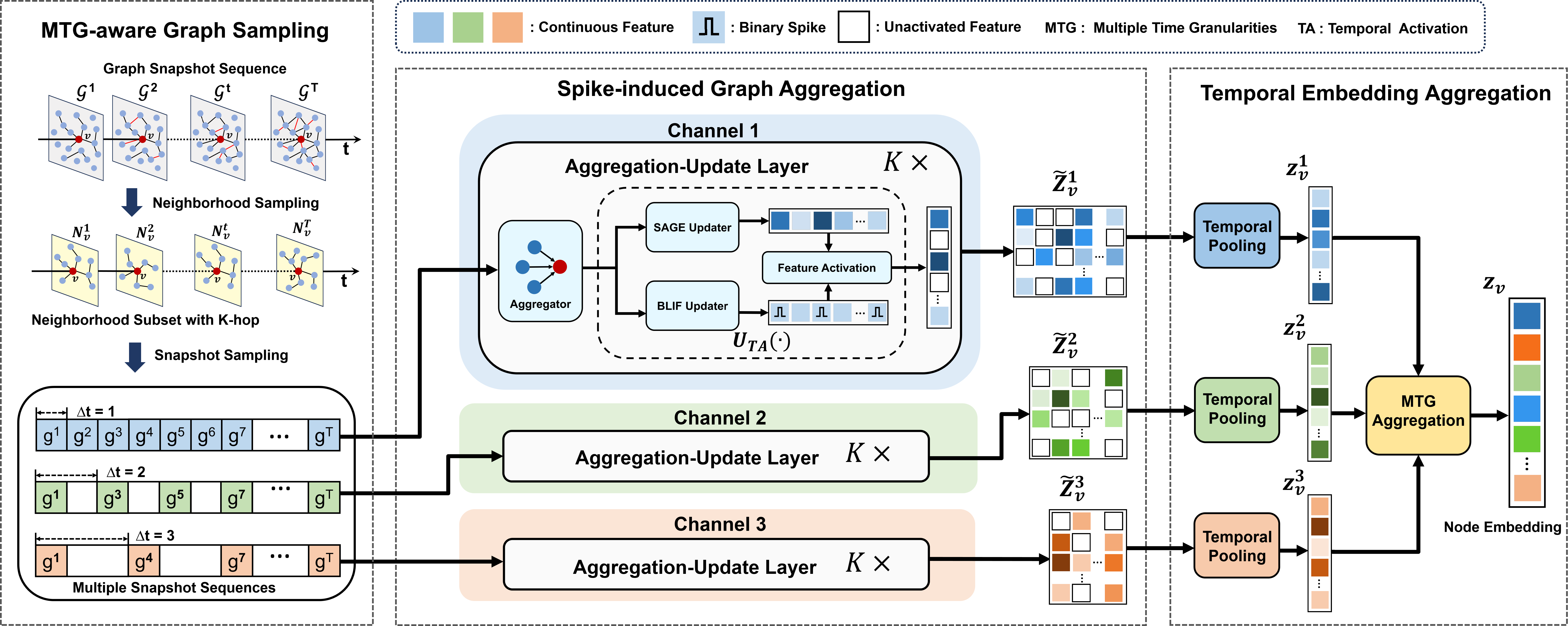}
  \caption{The framework of SiGNN. SiGNN consists of three fundamental steps: MTG-aware graph sampling , Spike-induced graph aggregation and temporal embedding aggregation. }
  \label{Framework}
\end{figure*}

\section{Method}\label{sec:OurMethod}

\subsection{Overview}
In this section, we introduce an overview of the proposed Spike-induced graph neural network (SiGNN) framework. As shown in Figure \ref{Framework}, SiGNN is composed of three primary steps aimed at learning spatial-temporal node representations in dynamic graphs.
\begin{itemize}
   \item MTG-aware Graph Sampling involves crucial neighborhood sampling and uniquely  incorporates simple snapshot sampling for dynamic graphs. This methodology facilitates the acquisition of snapshot sequences across Multiple Time Granularities (MTG). 
   \item Spike-induced Graph Aggregation employs independent channels for message aggregation on sampled snapshot sequences, integrating a novel Temporal Activation (TA) mechanism. Within TA mechanism, an advanced SNN neuron model called Bidirectional LIF (BLIF) is introduced to intricately capture the evolutionary dynamics of graphs. 
   \item Temporal Embedding Aggregation is designed to aggregate node embeddings from diverse time steps and granularities, thus forming intricate node representations tailored for downstream tasks.
\end{itemize}

\subsection{MTG-aware Graph Sampling}
MTG-aware Graph Sampling offers SiGNN profound insights into the structural pattern and evolutionary dynamics of graphs at various time scales.  In the context of a discrete dynamic graph, for a node $v$ at time $t$, its neighborhood subset $N^t_v$ is derived via neighborhood sampling from the graph $\mathcal{G}^t$. Herein, we adopt the temporal neighborhood sampling strategy as proposed by Li et al \cite{li2023scaling}. 

Execution of snapshot sampling comprises conducting successive rounds of sampling at varying intervals throughout the entire sequence of neighborhood subsets $\{N^1_v, N^2_v, N^3_v,..., N^T_v \}$, where $T$ denotes the total count of snapshots. With a chosen sampling interval  $\Delta t = 2$,  the resultant sequence $\{ N^1_v, N^3_v, N^5_v, ... , N^T_v \}$ emerges. This straightforward method facilitates gathering sequences of snapshots at distinct time granularities, crucial for SiGNN to discern the evolving dynamics of graph through multiple time-resolution lenses.

Significantly, MTG-aware graph sampling introduces extra snapshot sequences without notably increasing computational complexity. Sequences with coarser time granularities are substantially shorter than the complete sequence, maintaining a manageable total number of snapshots for the model to process.

\subsection{Spike-induced Graph Aggregation}
\subsubsection{Bidirectional LIF Neuron}
The traditional Leaky Integrate-and-Fire (LIF) model exhibits inhibitory responses to negative current inputs, complicating the generation of spike signals. While this characteristic mirrors that of biological neurons, it results in the loss of information regarding negative feature dynamics.  In contrast, the proposed BLIF model is designed to respond to both positive and negative inputs, thereby enhancing its capability to intricately extract the dynamics of node features within graphs.

As shown in Figure \ref{fig::blif}(a), the activity of a BLIF neuron at time $t$ unfolds across four distinct phases. Initially, in the \textbf{Integrate} phase, the neuron updates its membrane voltage to \( \tilde{V}^{t} \) by integrating the current input \( I^t \) with the previous membrane voltage \( V^{t-1} \). Subsequently, during the \textbf{Fire} phase, the neuron decides whether to emit a binary spike signal \( S^t \in \{0, 1 \}\) dependent on the previous threshold  \( V_{th}^{t-1} \). In the ensuing  \textbf{Reset} phase, the membrane voltage is adjusted to \( V^t \),  a modification determined by the spiking occurrence \( S^t \). Finally, by incorporating a adaptive \textbf{update} strategy, the threshold voltage is updated to \( V_{th}^t \) based on the \( V_{th}^{t-1} \) and \( S_t \). The specific computation scheme for each phase are as follows:
\begin{alignat}{2}\label{BLIF}
    &\textbf{Integrate : } &\tilde{V}^{t} &= V^{t-1} + \tau(V_{reset}- V^{t-1} + I^t)\\
    &\textbf{Fire : } &S^t &= F_b \left( \tilde{V}^{t}, V^{t-1}_{th} \right) \\
    &\textbf{Reset : } &V^t &= \tilde{V}^{t} (1 - S^t) + V_{reset} S^t\\
    &\textbf{Update : } &V_{th}^{t} &= \gamma V_{th}^{t-1} + (1 - \gamma ) S^t
\end{alignat}
where $\tau$ and $\gamma$ respectively represent the membrane voltage decay factor and the threshold voltage decay factor. Following the method of Fang et al.\cite{fang2021incorporating}, both factors are set as learnable parameters, thereby augmenting the expressivity of BLIF neurons. Notably, during the Fire phase, a novel bidirectional detection function symbolized as $\mathbf{F_b}(\cdot)$ is employed. The operational mechanism of this function is delineated as follow:
\begin{equation}\label{LIFi}
S^t = \Theta(\tilde{V}^{t} - V_{th}^{t-1}) + \Theta(-V_{th}^{t-1} - \tilde{V}^{t})
\end{equation}
where $\Theta(\cdot)$ represents the Heaviside step function, which is defined by $\Theta(x) = 1$ if $x \geq 0$ and 0 otherwise. Consequently, Thus, a spike signal is emitted either when the membrane voltage $\tilde{V}^{t}$ exceeds the positive threshold $V_{th}^{t-1}$ or falls below the negative threshold $-V_{th}^{t-1}$ as shown in Figure \ref{fig::blif}(b). This dual-threshold mechanism facilitates a more responsive firing behavior in the neuron model. Finally, the four computational phases of the BLIF neuron are encapsulated into a function as follow:

\begin{figure*}[t]	
        \centering
        \subfloat[]{
		\includegraphics[width=70mm]{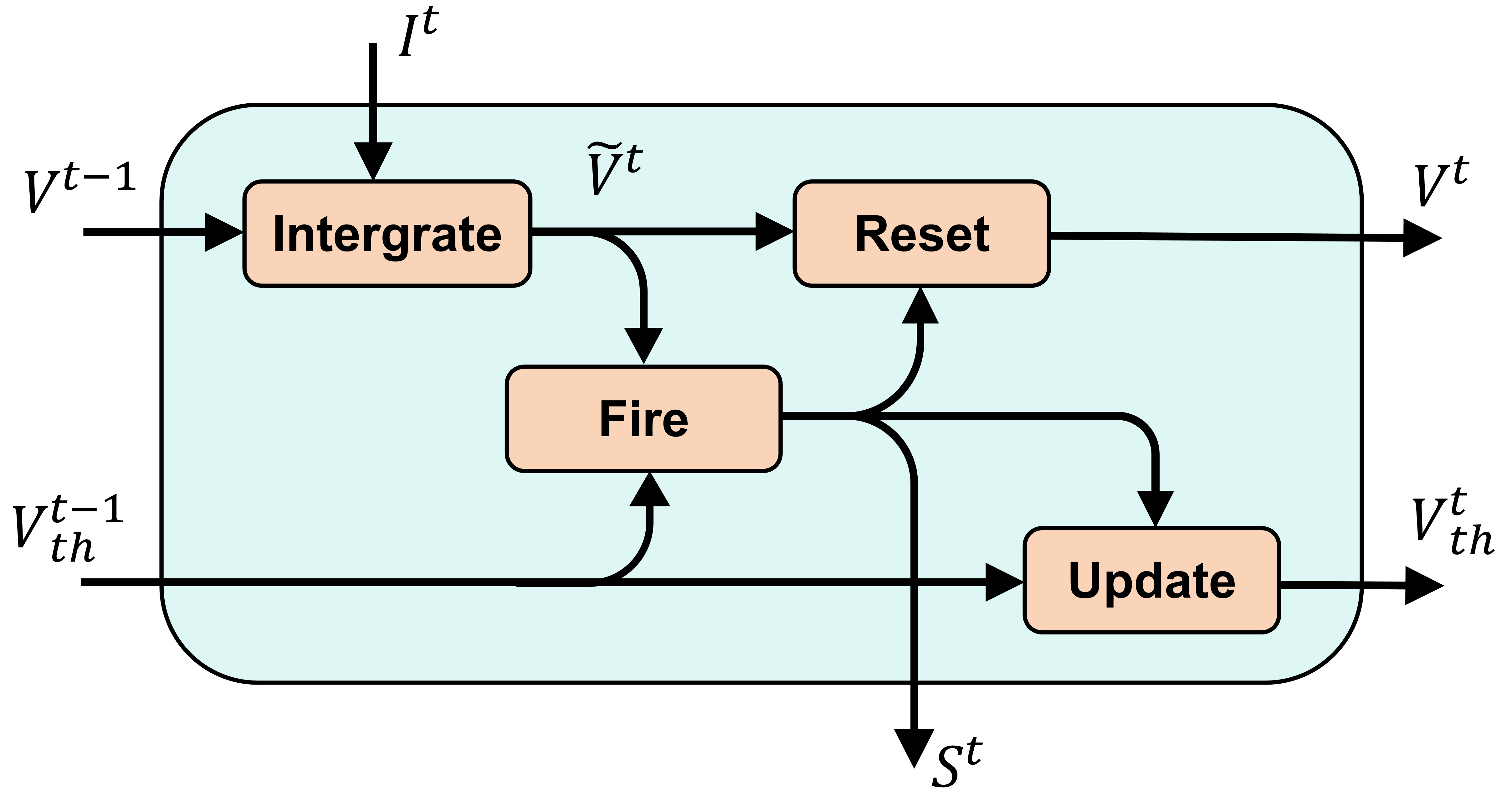}
        }
	   \subfloat[]{
		\includegraphics[width=50mm]{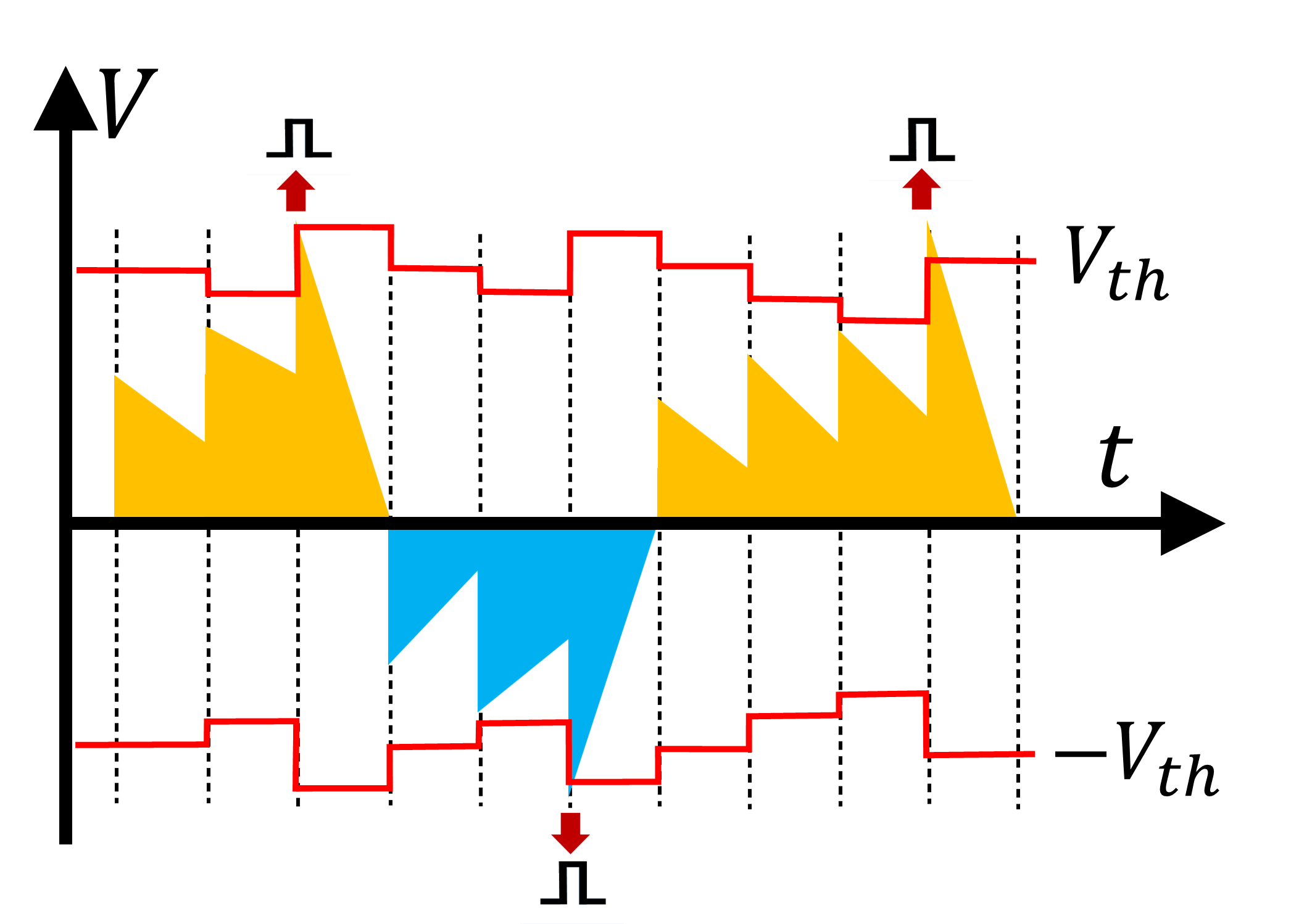} 
        }
	\caption{ The schematic diagram of the BLIF neuron model. Figure (a) illustrates the internal computational flow of BLIF neurons at time $t$, while Figure (b) represents the variation of membrane voltage over time and the bidirectional spike firing characteristics in BLIF neurons.}
\label{fig::blif}
\end{figure*}

\begin{equation}\label{LIFi}
S^t = BLIF \left( I^t; V^{t-1},V_{th}^{t}, V_{reset}, \tau,  \gamma \right)
\end{equation}
where $I^t$ and $S^t$ denote the input and output spikes of BLIF neurons, respectively, while $V^{t-1}$, $V_{th}^{t}$, $V_{reset}$, $\tau$ and  $\gamma$ represent the internal states and parameters governing neuron behavior.

\subsubsection{Temporal Activation}
Considering the limited representational capacity of binary spike signals, we introduce the TA mechanism for leveraging the SNNs' temporal dynamics in a effective and harmonious manner. Unlike explicit spike-based methods, TA mechanism implicitly employs spikes as activation signals to temporally control feature propagation in GNNs. 

In the aggregation-update layer, SiGNN initiates by aggregating messages from neighboring nodes, a pivotal step that empowers nodes to distill insights pertaining to their immediate local environments. Notably, SiGNN distinctively harnesses the proposed TA mechanism to update node representations. The procedure for each aggregation-update layer unfolds as follows: 
\begin{align}
&h^{t, (k)}_{N_v} = Agg \left( h^{t, (k)}_u, \forall u \in N^t_v \right) \\
&h^{t, (k+1)}_v = U_{TA} \left( h^{t, (k)}_v,  h^{t, (k)}_{N_v} \right)
\end{align}
where $h^{t, (k)}_v$ denotes the hidden representation of node $v$ at $k_{th}$ layer and time step $t$. Specifically, the initial representation $h^{t, (0)}_v$ is set to $x^t_v \in \mathbf{X^t}$ and the acquired node embedding $h^{t, (K)}_v$ after $K$ layers. $AGG(\cdot)$ is the aggregation function, which usually is a permutation-invariant function such as mean, sum or max. The function $U_{TA}(\cdot)$ employs the TA mechanism to update node representation in the following manner:
\begin{align}
&\tilde{h}^{t, (k+1)}_v = \mathbf{\sigma} \left( h^{t, (k)}_{N_v} \mathbf{W}^{(k)}_{h} + h^{t, (k)}_v \mathbf{B}^{(k)}_{h} \right) \\
&{s}^{t, (k+1)}_v = BLIF \left( h^{t, (k)}_{N_v} \mathbf{W}^{(k)}_{s} + h^{t, (k)}_v \mathbf{B}^{(k)}_{s} \right) 
\end{align}
where $\mathbf{W}^{(k)}_{h}$, $\mathbf{B}^{(k)}_{h}$, $\mathbf{W}^{(k)}_{s}$, and $\mathbf{B}^{(k)}_{s}$ represent trainable parameters at the $k_{th}$ layer. The TA mechanism initially employs two pathways for integrating neighborhood representations $h^{t, (k)}_{N_v}$ and self-representations $h^{t, (k)}_v$. Subsequently, the integrated results are separately fed into the sigmoid function $\mathbf{\sigma}$ and the aforementioned $BLIF(\cdot)$ neurons. Therefore, a node representation $\tilde{h}^{t, (k+1)}_v$ with real-valued features and a binary spike vector ${s}^{t, (k+1)}_v$ is obtained. In order to amalgamate information from these two sources, the TA mechanism employs the ensuing operation:
 \begin{equation}\label{LIFi}
h^{t, (k+1)}_v = {s}^{t, (k+1)}_v \odot \tilde{h}^{t, (k+1)}_v
\end{equation}
where $\odot$ is element-wise multiplication that indicates the utilization of spike signals to activate hidden features. After this simple operation, the features within acquired representation $h^{t, (k+1)}_v$ do not maintain consistency with $\tilde{h}^{t, (k+1)}_v$ at every time step. Instead, these features become visible solely upon activation by spike signals, with any unactivated features maintaining a value of zero.

Essentially, spikes in TA mechanism play a crucial role in controlling the propagation of features along the temporal dimension in GNNs. After aggregating all snapshots from the dynamic graph, a temporal-aware node embedding matrix $\mathbf{\tilde{Z}}^1_v \in \mathbb{R}^{d \times T}$ (assuming the snapshot sampling interval is 1) is derived by stacking embeddings sequence $\{h^{1, (K)}_v, h^{2, (K)}_v, ..., h^{T, (K)}_v\}$. Owing to the effect of the TA mechanism,  the temporal distribution of  features in $\mathbf{\tilde{Z}}^1_v$ implicitly reveals the evolution dynamics of nodes. Thus, $\mathbf{\tilde{Z}}^1_v$ encompasses the spatial-temporal information of nodes and effectively mitigates embeddings' feature redundancy across distinct time steps. Similarly,  the node embedding matrices ($\mathbf{\tilde{Z}}^2_v \in \mathbb{R}^{d \times \frac{T}{2}}$, $\mathbf{\tilde{Z}}^3_v \in \mathbb{R}^{d \times \frac{T}{3}}, ...)$ at other time granularities is acquired.

\subsection{Temporal Embedding Aggregation}
Given the unique characteristics of the node embedding matrix from spike-induced graph aggregation, it is imperative to consider the temporal positioning of features when converting this matrix into a low-dimensional embedding.  For instance with $\tilde{Z}^1_v$, we employ the following technique for temporal pooling:
 \begin{equation}\label{LIFi}
z^1_v =  ( \mathbf{\tilde{Z}}^1_v \odot \mathbf{P}_{1} ) \cdot \mathbf{1}
\end{equation}
where $\mathbf{P}_{1} \in \mathbb{R}^{d \times T}$ is a trainable parameter matrix that imparts temporal distinctiveness when aggregating features across different time steps. The vector $\mathbf{1}$ containing $T$ ones, transforms the node embedding matrix into a vector $z^1_v \in \mathbb{R}^{d}$, which represents the acquired node embedding at the respective time granularity channel.

Through temporal pooling, node embeddings at various time granularities $(z^1_v, z^2_v, z^3_v,...)$ are aggregated. MTG aggregation is employed to derive the final node representation $z_v$. Herein, there are various strategies available for aggregating embeddings across multiple time granularities,  including methods such as averaging, concatenation, attention mechanism, etc.


\section{Experiments}\label{sec:experiment}
In this section, we conducted node classification experiments on real-world dynamic graph datasets to demonstrate the outstanding performance of our SiGNN.

\subsection{Datasets}
Three real-world dynamic graph datasets from varied domains are used to evaluate our model. The DBLP dataset is an academic co-authorship network, capturing collaborations in scholarly publications. The Tmall \cite{lu2019temporal} dataset, derived from e-commerce sales data, forms a bipartite graph representing consumer-product interactions. The Patent 
 \cite{hall2001nber} dataset is a citation network among U.S. patents, illustrating the knowledge flow in patent innovations.  Detailed statistics of these datasets are presented in Table \ref{dataset}. In specific experiments, in order to reduce memory and computational resource consumption, we compressed the time steps of the Tmall and Patent datasets to 19 steps and 13 steps, respectively.

\begin{table}

  \centering
  \caption{Statistics of the Datasets}
  \begin{tabular}{@{}l c c c@{}}
    \toprule
    \multicolumn{1}{c}{\textbf{Attributes}} & \multicolumn{1}{c}{\textbf{DBLP}} & \multicolumn{1}{c}{\textbf{Tmall}} & \multicolumn{1}{c}{\textbf{Patent}} \\
    \midrule
    Nodes & 28,085 & 577,314 & 2,738,012 \\
    Edges & 236,894 & 4,807,545 & 13,960,811 \\
    Classes & 10 & 5 & 6 \\
    Time steps & 27 & 186 & 25 \\
    \bottomrule
  \end{tabular}
  \label{dataset}
\end{table}

\begin{table*}[h]
  \footnotesize
  \centering
  \caption{Performance of SiGNN and the baselines on the task of temporal node classification (\%).}
  \label{table_5}
\resizebox{\textwidth}{!}
{
\begin{tabular}{c c c c c |c c c|c c c}
    \toprule
    \textbf{Dataset} & \textbf{Metrics} & \textbf{Training}  & \textbf{DeepWalk} & \textbf{Node2Vec} & \textbf{JODIE}  & \textbf{EvolveGCN} & \textbf{TGAT} & \textbf{SpikeNet} & \textbf{SiGNN} \\
    \midrule
    \multirow{6}{*}{\textbf{DBLP}} & \multirow{3}{*}{\textbf{Macro-F1}} &  \textbf{40\%} & 67.08 & 66.07 & 66.73±1.0 &67.22±0.3 & 71.18±0.4 & {70.71±0.38} & \textbf{74.66±0.26} \\
    & & \textbf{60\%} & 67.17 & 66.81 & 67.32±1.1 & 69.78±0.8 & 71.74±0.5 & {72.13±0.22} & \textbf{76.68±0.36} & \\
    & & \textbf{80\%} & 67.12 & 66.93 & 67.53±1.3 & 71.20±0.7 & 72.15±0.3 & {73.91±0.49} & \textbf{78.74±0.22}   \\
    \cmidrule{2-11}
    & \multirow{3}{*}{\textbf{Micro-F1}} & \textbf{40\%} & 66.53 & 66.80  & 68.44±0.6 & 69.12±0.8 &71.10±0.2 & {72.00±0.27} & \textbf{75.36±0.18}\\
    & & \textbf{60\%} & 66.89  & 67.37 & 68.51±0.8 & 70.43±0.6 & 71.85±0.4 & {73.21±0.31} & \textbf{77.02±0.20} \\
    & & \textbf{80\%} & 66.38 & 67.31  &68.80±0.9 & 71.32±0.5  & 73.12±0.3 & {74.53±0.50} & \textbf{79.16±0.21} \\
    \midrule
    \multirow{6}{*}{\textbf{Tmall}} & \multirow{3}{*}{\textbf{Macro-F1}} &  \textbf{40\%} & 49.09 & 54.37 & 52.62±0.8 & 53.02±0.7 & 56.90±0.6 & {59.38±0.12} & \textbf{62.47±0.20} \\
    & & \textbf{60\%} & 49.29 & 54.55& 54.02±0.6 & 54.99±0.7 & 57.61±0.7 & {61.14±0.12} & \textbf{64.40±0.31} \\
    & & \textbf{80\%} & 49.53 & 54.58 & 54.17±0.2 & 55.78±0.6 & 58.01±0.7 & {62.89±0.34} & \textbf{66.14±0.28}\\
    \cmidrule{2-11}
    & \multirow{3}{*}{\textbf{Micro-F1}} & \textbf{40\%} & 57.11 & 60.41 & 58.36±0.5 & 59.96±0.7 & 62.05±0.5 & {63.50±0.13} & \textbf{66.43±0.07} \\
    & & \textbf{60\%} & 57.34 & 60.56 & 60.28±0.3 & 61.19±0.6 & 62.92±0.4 & {65.03±0.14} & \textbf{68.15±0.14}\\
    & & \textbf{80\%} & 57.88 & 60.66  & 60.49±0.3 & 61.77±0.6  & 63.32±0.7. & {66.53±0.33} & \textbf{69.63±0.08}\\
    \midrule
    \multirow{6}{*}{\textbf{Patent}} & \multirow{3}{*}{\textbf{Macro-F1}} &  \textbf{40\%} & 72.32±0.9 & 69.01±0.9 & 77.57±0.8 & 79.67±0.4 & 81.51±0.4 & {83.53±0.6} & \textbf{84.45±0.04}\\
    & & \textbf{60\%} & 72.25±1.2 & 69.08±0.9 & 77.69±0.6 & 79.76±0.5 & 81.56±0.6 & {83.85±0.7} & \textbf{84.81±0.05} \\
    & & \textbf{80\%} & 72.05±1.1  & 68.99±1.0 & 77.67±0.4 & 80.13±0.4 & 81.57±0.5 & {83.90±0.6} & \textbf{84.91±0.03}\\
    \cmidrule{2-11}
    & \multirow{3}{*}{\textbf{Micro-F1}} & \textbf{40\%} & 71.57±1.3 & 68.14±0.9 & 77.64±0.7 & 79.39±0.5 & 80.79±0.7 & {83.48±0.8} & \textbf{84.41±0.02}\\
    & & \textbf{60\%} & 71.53±1.0  & 68.20±0.7 & 77.89±0.5 & 79.75±0.3 & 80.81±0.6 & {83.80±0.7} & \textbf{84.78±0.04} \\
    & & \textbf{80\%} & 71.38±1.2 & 68.10±0.5  & 77.93±0.4 & 80.01±0.3  & 80.93±0.6 & {83.88±0.9} & \textbf{84.89±0.03} \\
    \bottomrule
  \end{tabular}
}
\end{table*}

\subsection{Overall Performances}
Our experiments primarily delve into the evaluation of node representations in temporal node classification. This task utilizes the complete graph structure at each snapshots for representation learning on dynamic graphs. 

We conducted meticulous comparative experiments to assess our SiGNN against existing methods. The methods include static graph representation learning algorithms (DeepWalk and Node2Vec), as well as typical dynamic graph methods (JODIE, EvolveGCN, and TGAT), which employ RNNs or attention mechanisms to capture temporal dynamics. Our evaluation also encompasses SpikeNet, a pioneering method based on SNNs. Following Lu et al's methodology \cite{lu2019temporal}, the ratio of training set is set as 40\%, 60\%, and 80\%. We report the results in terms of Macro-F1 and Micro-F1 in Table \ref{table_5}. For three datasets, we adopted the results from Li et al.\cite{li2023scaling} for all methods excluding SpikeNet.

As evident from the results, the proposed SiGNN consistently outperforms all baseline methods in all cases. Specifically, SiGNN outperforms the strongest baseline by a significant margin, achieving a 4.83\% improvement on DBLP and 3.25\% on Tmall in the best-case scenario, along with an approximate 1.0\% enhancement on the Patent dataset. The methods designed for dynamic graphs consistently outperform the static graph methods. This observation highlights the crucial role of incorporating temporal information to achieve more expressive node representations. In addition, SpikeNet and SiGNN outperform other state-of-the-art methods including EvolveGCN, JODIE, and TGAT, which use different temporal models. This superior performance demonstrates the effectiveness of SNNs in temporal processing, showcasing their potential in the realm of dynamic graph representation learning.

Among SNN-based methods, our SiGNN notably stands out, achieving significant improvements. On three real-world datasets, it consistently surpasses the pioneering SpikeNet in all scenarios. This observation underscores the superior utilization of SNN's temporal dynamics in SiGNN, particularly through its adoption of the Temporal Activation (TA) mechanism. Overall, SiGNN excels in learning high-quality spatial-temporal node representations, exhibiting outstanding performance in temporal node classification tasks.


\subsection{Ablation Study}
To verify the utilization of the temporal activation mechanism and introducing multiple time granularity do contributes to enhancing the quality of the learned node representations, we conduct several ablation experiments.
\subsubsection{Temporal Activation Mechanism}
As one of our contributions, the Temporal Activation (TA) mechanism effectively incorporates temporal information into node representations. To provide empirical evidence, we compared experimental results using the TA mechanism with LIF or BLIF models, versus not using the TA mechanism (TA-LIF, TA-BLIF and without TA), while keeping other hyperparameters consistent.

\begin{figure}
  \centering
  \includegraphics[width=13cm]{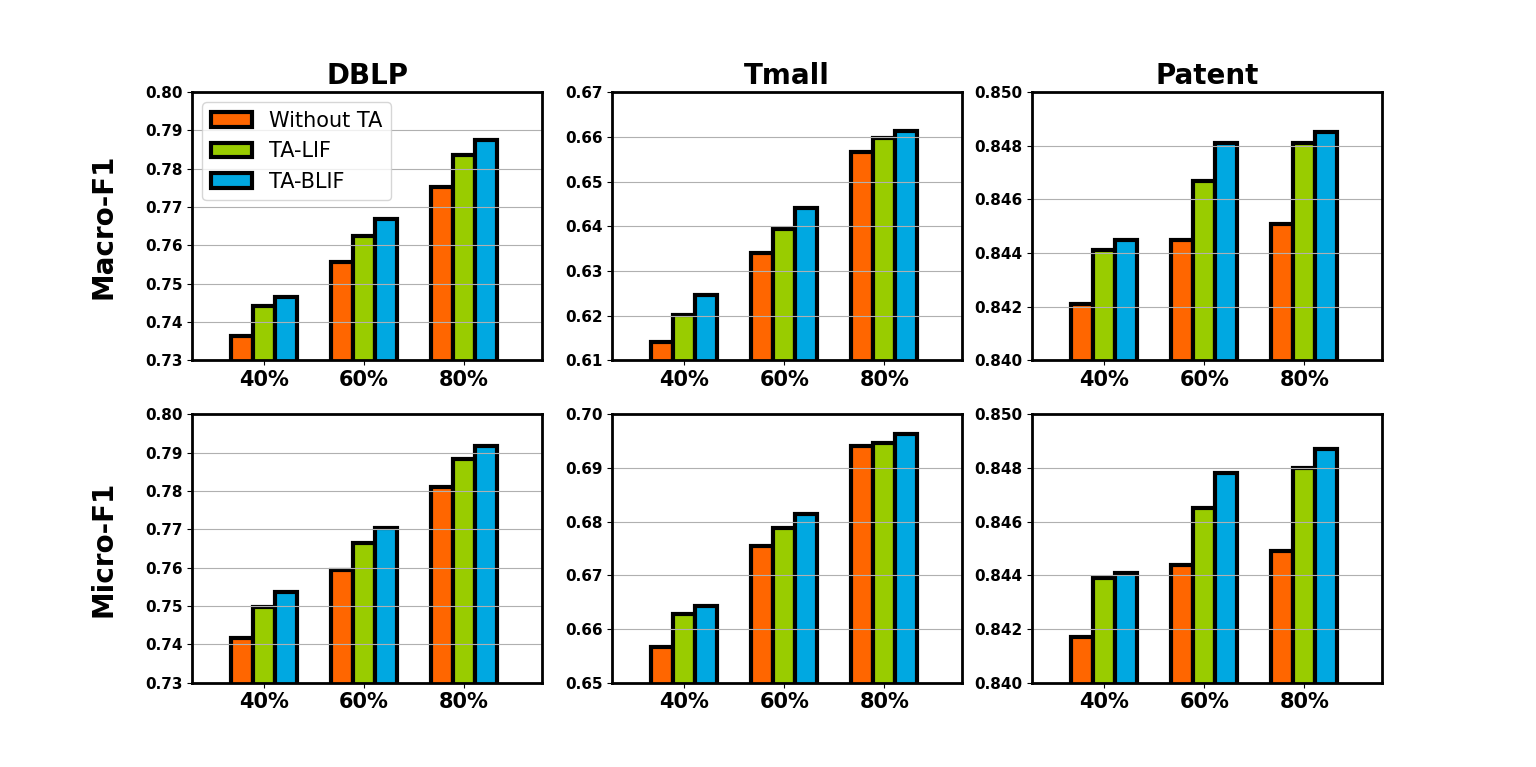}
  \caption{The Macro-F1 and Micro-F1 metrics for three methods: TA-BLIF, TA-LIF, and without TA mechanism. The x-axis represents various training ratios.}\label{TA}
\end{figure}

The Micro-F1 and Macro-F1 metric results of the three methods is presented in Figure \ref{TA}. It's evident that adopting a temporal activation mechanism enhances the model's performance, underscoring its effectiveness in the temporal processing of dynamic graph. Notably, on the DBLP dataset, employing the TA mechanism with the BLIF model improved performance by up to 1.35\% over methods without the TA mechanism. In addition, utilizing the TA mechanism with the BLIF model yields superior results compared to the LIF model, achieving an improvement of around 0.5\% in the best-case scenario. This demonstrates the superior dynamic capturing ability and expressiveness of the proposed BLIF model in tasks related to dynamic graph representation learning.

\subsubsection{Multiple Time Granularities}
SiGNN harnesses Multiple Time Granularities (MTG) to holistically comprehend the temporal dynamics of dynamic graphs from a multi-scale vantage point. To probe the impact of introducing varying numbers of time granularities on model performance, experiments were conducted across a spectrum ranging from one to five granularities. These experiments were conducted on three datasets, maintaining a consistent training ratio of 60\%, while upholding homogeneity in other hyperparameters.

\begin{figure}
  \centering
  \includegraphics[width=13cm]{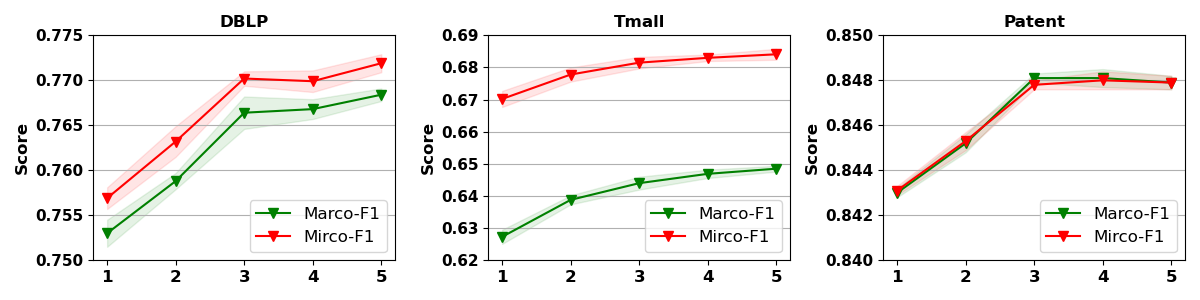}
  \caption{The performance of models with varying numbers of time granularity channels. The x-axis represents the count of introduced time granularities.}\label{mtg}
\end{figure}

The Micro-F1 and Macro-F1 metric results are presented in Figure \ref{mtg}. As the number of time granularity introduced increases, the model's performance improves. For instance, the performance of the 3-channel model exceeds that of the single-channel model by 1.33\% and 1.67\% on DBLP and Tmall datasets, respectively, which clearly demonstrates the benefits of incorporating multi-scale temporal information in enhancing the quality of learned representations. Furthermore, when the number of introduced time granularities exceeds 3, the performance improvement of the model becomes less pronounced, especially for the DBLP dataset. This observation indicates that the gain from incorporating coarser time granularity information is limited.

\subsection{Insights of Spike Signals}
\begin{figure}
  \centering
  \includegraphics[width=13cm]{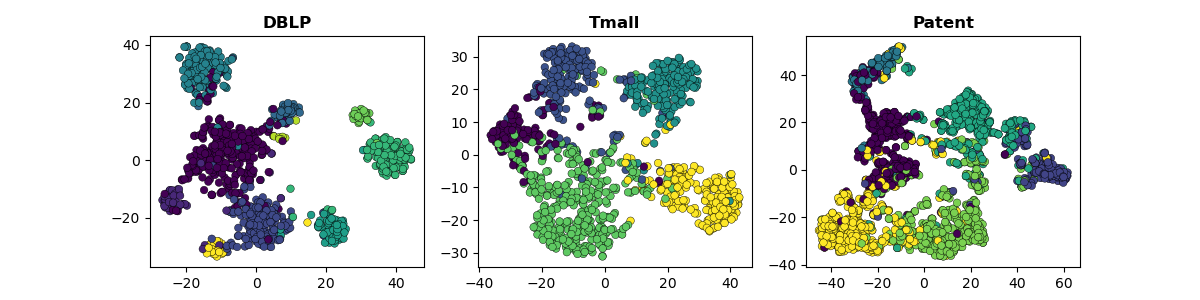}
  \caption{Visualization of node embeddings derived from spike signals. The visualization results rely on the t-SNE technique to map high-dimensional spike signals into a two-dimensional space.}\label{ts}
\end{figure}

The temporal distribution of spikes provides valuable insights for understanding the evolution patterns of dynamic graphs, as the timing of spike signal occurrences carries significant information. Therefore, we explored the capability of spikes to characterize node features. Accordingly, we directly employ the spike signals generated by SNN neurons as node embeddings for node classification task. Figure \ref{ts} illustrates the visualization of spike-based node representations within the embedding space. 

Remarkably, even with the simplicity of binary spike signals, there is a noticeable separation among nodes from different classes in the embedding space. This observation underscores the efficacy of utilizing temporal dynamics encoded in spike signals in dynamic graph analysis. Hence, SiGNN employs the temporal activation mechanism, which effectively incorporates the information embedded in the temporal distribution of spike signals.  

\subsection{Voltage Decay Rate $\tau$ and Threshold Decay Rate $\gamma$}
The membrane voltage decay rate and threshold voltage decay rate are key parameters of the BLIF model. Table \ref{tau} comprehensively examines the values of learned $\tau$ and $\gamma$ across different datasets and time granularity channels. Here, $\tau_1$, $\tau_2$, and $\tau_3$ represent the average membrane voltage decay rates of all BLIF neurons across three different time granularity channels, while $\gamma_1$, $\gamma_2$, and $\gamma_3$ denote the average threshold voltage decay rates of all BLIF neurons similarly.

\begin{table}
  \centering
  \caption{The average membrane voltage decay rates ($\tau$) and threshold decay rate ($\gamma$) of BLIF neurons across different datasets and time granularity channels.}
  \begin{tabular}{@{}l c c c| c c c@{}}
    \toprule
    \multicolumn{1}{c}{\textbf{Datasets}} & \multicolumn{1}{c}{\textbf{$\tau_1$}} & \multicolumn{1}{c}{\textbf{$\tau_2$}} & \multicolumn{1}{c}{\textbf{$\tau_3$}} &\multicolumn{1}{c}{\textbf{$\gamma_1$}} &\multicolumn{1}{c}{\textbf{$\gamma_2$}}&\multicolumn{1}{c}{\textbf{$\gamma_3$}}\\ 
    \midrule
    DBLP & 0.554  &0.570 &  0.579 & 0.763  &0.691 &  0.729 \\
    Tmall & 0.651 &0.661 & 0.649 & 0.581  &0.459 &  0.427\\
    Patent & 0.797 & 0.783 & 0.742 & 0.406 &0.365 &  0.341 \\
    \bottomrule
  \end{tabular}
  \label{tau}
\end{table}

Although slight variations in the membrane voltage decay rates are observed within the same dataset across different time granularity channels, significant disparities emerge across different datasets. This highlights the SNN's ability to adjust its dynamic parameters based on the characteristics of specific datasets, optimizing adaptability to various input patterns. Notably, for the DBLP dataset, neurons exhibit smaller $\tau$, approximately 0.56, indicating a higher responsiveness of BLIF neurons to inputs and a preference for capturing rapidly changing features. Additionally, BLIF neurons exhibit larger $\gamma$, approximately 0.72, suggesting a higher propensity for spike firing. In contrast, for the Patent dataset, neurons features larger $\tau$ and smaller $\gamma$, implying neurons respond more weakly to inputs, focusing on capturing sustained dynamic features.

\begin{table*}[h]
  \footnotesize
  \centering
  \caption{The performance of SiGNN with various aggregation strategies for multiple time granularities on the task of temporal node classification.}
  \label{mtga}
  \resizebox{\textwidth}{!}{
    \begin{tabular}{ccccccc}
      \toprule
      \textbf{Dataset} & \textbf{Metrics} & \textbf{Training}  & \textbf{Average} & \textbf{Max} & \textbf{Concat}  & \textbf{Attention}\\
      \midrule
      \multirow{6}{*}{\textbf{DBLP}} & \multirow{3}{*}{\textbf{Macro-F1}} 
      &  \textbf{40\%} & 74.66 & 74.30 & \textbf{74.74} &74.45 \\
      & & \textbf{60\%} & \textbf{76.68} & 76.20 & 76.65 & 76.04 \\
      & & \textbf{80\%} & 78.74 & 78.25 & \textbf{78.96} & 78.47 \\
      \cmidrule{2-7}
      & \multirow{3}{*}{\textbf{Micro-F1}} 
      & \textbf{40\%} & \textbf{75.36} & 75.01  & 75.31 & 75.24\\
      & & \textbf{60\%} & \textbf{77.02}  & 76.54 & 76.93 & 76.62\\
      & & \textbf{80\%} & \textbf{79.16} & 78.83  &79.11 & 79.02 \\
      \midrule
      \multirow{6}{*}{\textbf{Tmall}} & \multirow{3}{*}{\textbf{Macro-F1}} &  \textbf{40\%} & \textbf{62.47} & 61.96 & 62.46 &62.31 \\
      & & \textbf{60\%} & \textbf{64.40} & 63.33 & 64.16 & 64.35 \\
      & & \textbf{80\%} & \textbf{66.14} & 65.11 & 65.91 & 66.07 \\
      \cmidrule{2-7}
      & \multirow{3}{*}{\textbf{Micro-F1}} 
      & \textbf{40\%} & 66.43 & 65.69  & 66.46& \textbf{66.47}\\
      & & \textbf{60\%} & \textbf{68.15}  & 67.51 & 67.88 & 68.12\\
      & & \textbf{80\%} & 69.63 & 69.11  &69.55 & \textbf{69.65} \\
      \bottomrule
    \end{tabular}
  }
\end{table*}
\subsection{Aggregation Strategy for MTG}
The analysis of dynamic graphs across multiple time granularities offers a unique perspective for understanding the evolution of graph structures, yet effectively integrating embeddings across these varied granularities is worth considering. In this experiment, we compare four aggregation methods: \textbf{Max}, \textbf{Average}, \textbf{Concat}enation, and an \textbf{Attention} mechanism-based aggregation, aiming to uncover the potential influence of different aggregation strategies on enhancing the quality of dynamic graph representations. During the experiment, the number of time granularity channels was set to three and all other hyperparameters were kept consistent.

As illustrated in Table \ref{mtga}, it was found that models employing various aggregation methods yielded similar performances on DBLP and Tmall datasets. Notably, average aggregation methods generally surpassed others in most scenarios. Average aggregation uniformly integrates information from various time granularities, promoting a balanced fusion that improves the understanding of dynamic graphs' temporal nuances. Although attention-based aggregation offers adaptability in valuing different temporal information, it requires ample data and precise tuning. In contrast, the simplicity of average aggregation might be more effective when dealing with limited data or complex models.

\subsection{Spike Firing Rate and Graph Dynamics}
\begin{figure}
  \centering
  \includegraphics[width=13cm]{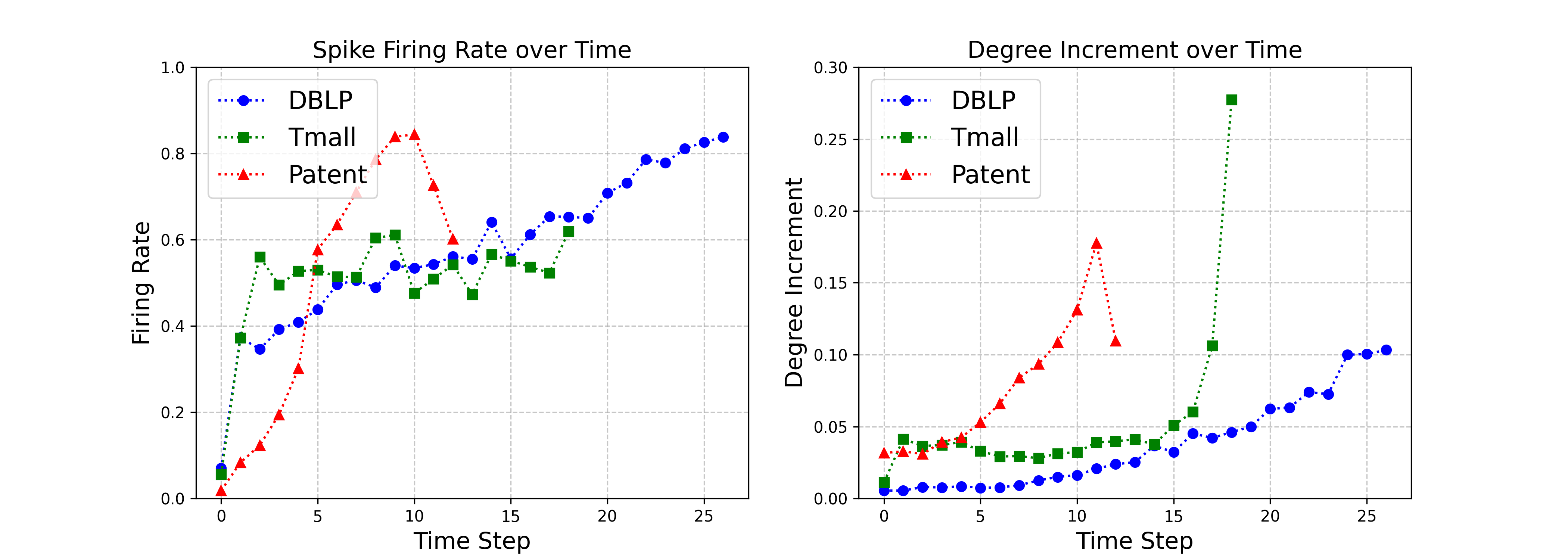}
  \caption{Average spike firing rates and graph degree increment at each time step. The spike firing rate is calculated by: (number of spikes fired / number of BLIF neurons) and the degree increment is calculated by: (degree increase /  total degree number). }\label{sr}
\end{figure}

To delve deeper into the interplay between the temporal dynamics of SNN neurons and the evolution of dynamic graphs, we conducted comprehensive experiments. Our investigation focused on examining the correlation between the average spike firing rate of BLIF neurons and the degree increment in the dynamic graph at each time step. The results revealed a notable alignment in their variations, as illustrated in Figure \ref{sr}.

In the DBLP dataset, the gradual increase in node degrees corresponds to a similar trend observed in the average spike firing rate of SNN neurons. Likewise, in the Patent dataset, the initial rise followed by a subsequent decline in node degrees is mirrored by corresponding fluctuations in the spike firing rate, initially peaking at 0.8 before gradually decreasing to approximately 0.6 in later time steps. This suggests a potential correlation between the activity level of SNN neurons and the evolutionary degree of dynamic graphs. This consistency implies that SNNs are capable of adapting to changes in graph topological structures, tending to become more active in response to significant alterations in dynamic graphs. Consequently, the effectiveness of SNNs as a temporal architecture in learning dynamic graph representations is underscored, highlighting their capability to capture the spatiotemporal evolution patterns of dynamic graphs.

\section{Conclusions and Future Work}\label{sec:Conclusion}
In this paper, we propose an innovative framework named SiGNN for learning node representations on dynamic graphs. SiGNN strategically  leverages  SNN as a temporal processing model and incorporates multi-temporal granularity analysis of dynamic graphs to capture spatiotemporal features across multiscale. In SiGNN, a novel temporal activation mechanism is introduced to effectively incorporate the temporal dynamics of SNN into the static GNN model while maintaining high model performance. Extensive experiments validate the superior performance of SiGNN in learning dynamic graph representations, surpassing existing SNN-based methodologies. Furthermore, we conduct an interpretative exploration of the dynamic characteristics of SNNs, revealing their seamless integration within the realm of dynamic graph representation learning.

Significantly, the proposed SiGNN exhibits scalability, as its approach of harnessing the temporal dynamics of SNN is adaptable to various static GNNs. In this work, GraphSAGE is exclusively utilized as the static GNN architecture for elucidation and experimentation. Going forward, we intend to investigate the integration of spike-induced temporal methods with other GNN models. This exploration aims to uncover the potential of SNN in the domain of dynamic graph representation learning and potentially extend to addressing other temporal data processing challenges.



\bibliography{main}

\bibliographystyle{unsrt}





\end{document}